\documentclass{article}
\usepackage{spconf,amsmath,graphicx}
\usepackage{amssymb}
\usepackage{booktabs} 
\usepackage{color}
\usepackage{hyperref}

\title{Sparse Generation: Making Pseudo Labels Sparse for
Point Weakly Supervised Object Detection on Low Data Volume}
\name{Chuyang Shang$^1$$^\dagger$, Tian Ma$^1$$^\dagger$, Wanzhu Ren$^1$, Yuancheng Li$^1$, Jiayi Yang$^1$ 
\thanks{This work was supported by the National Key R\&D program of China (2022ZD0119005), and National Key Laboratory of Spatial Intelligent Control Technology (HTKJ2024KL502027).}
}
\address{$^1$Xi'an University of Science and Technology, Xi'an, China} 

\begin{document}
%
\maketitle
\begin{abstract}
Existing pseudo label generation methods for point weakly supervised object detection are inadequate in low data volume and dense object detection tasks. We consider the generation of weakly supervised pseudo labels as the model’s sparse output, and propose Sparse Generation as a solution to make pseudo labels sparse. The method employs three processing stages (Mapping, Mask, Regression), constructs dense tensors through the relationship between data and detector model, optimizes three of its parameters, and obtains a sparse tensor, thereby indirectly obtaining higher quality pseudo labels, and addresses the model’s density problem on low data volume. Additionally, we propose perspective-based matching, which provides more rational pseudo boxes for prediction missed on instances. In comparison to the SOTA method, on four datasets (MS COCO-val, RSOD, SIMD, Bullet-Hole), the experimental results demonstrated a significant advantage. \textcolor{blue}{https://github.com/Trumpetertimes/Sparse-Generation} 
\end{abstract}
\begin{keywords}
computer vision, object detection, weakly supervised object detection, semi-supervised object detection 
\end{keywords}
\section{Introduction}
\label{sec:intro}

In recent years, methods based on PWSOD (Point Weakly Supervised Object Detection) \cite{chen2021points,chen2022point} aroused research interests in academia. It only needs to annotate a very small amount of supervised annotation data, other data can use the low-cost weakly supervised annotation format, which can greatly reduce the workload \cite{fu2023ufo2,ren2020ufo}.
Compared with semi-supervised object detection(SSOD) \cite{sohn2020simple,sohn2020fixmatch,tarvainen2017mean,zhou2022dense} methods, utilizing these weakly supervised annotations information could better guide the model training. 

The output process of the object detector currently using CNN \cite{long2015fully} as the backbone, can be viewed as a region selection process from dense to sparse \cite{sun2021sparse}.
The model’s backbone network usually contains thousands of region proposals, which will be filtered out under the selection of the detection head.

\textbf{However}, in weakly supervised methods, due to the inability to directly obtain the bounding box from weakly supervised annotation data, it is often necessary to rely on additional networks or detectors themselves to assist in generating labels. For the limited use of supervised annotation data, the detection head degrades its functionality. Inability to obtain accurate instance areas, also the number of pseudo labels \cite{lee2013pseudo} generated greatly exceeds the number of instances in detected images, which is particularly prominent in low data volume and dense instances detection tasks.

\textbf{Disadvantages of using network} Prior works used additional networks \cite{chen2022point,zeng2019wsod2,tang2017multiple,cheng2020high,zhang2022group} 
specifically for pseudo labels generation, we observe their networks output would be a dense set in low training data volume, and prone to localized focusing problem \cite{zeng2019wsod2,tang2017multiple,cheng2020high}, this issue is usually characterized by overlapping boxes and missed prediction on instance. Repeated optimization in these imprecise subsets may not be able to tap into more potential of the algorithm. Moreover, the utilization of supplementary networks is kind of superfluous, with a significant expenditure of resources and time.

\textbf{Lack of direct regression to pseudo labels} 
Some previous works \cite{sohn2020simple, zhang2022group, ge2023point} employed matching/filtering mechanisms to select or assign pseudo labels. However, these approaches lack directly learning and regression for pseudo labels, and do not fully account for the global information present in all prediction results. Although the dense prediction results on instances are mostly inaccurate, all these pseudo boxes represent to some extent the decision tendency from the detection network. We believe that  mapping these pseudo boxes into the feature space, and regress them on a relatively limited volume of data could generate a more reasonable prediction of the pseudo labels.

\textbf{In this paper}, we propose Sparse Generation starting from the root problems that constraining the performance of weakly supervised object detection in low data volume, that is, to avoid the dense prediction result, and localized focusing. Sparse Generation employs a non-network approach, the mapping stage effectively obtains the representation of dense pseudo boxes with spatial dimension; the mask stage constructs a dense tensor map to cover the whole instance as much as possible in order to reduce the localized focusing; the regression stage calculates the bounding box location from dense distributions, and refines the final pseudo boxes, thereby obtaining the sparse output. At the same time, the non-networked approach reduces consumption of computing resources. We also provide a new dataset Bullet-Hole to evaluate the performance for dense objects.

\section{Sparse Generation}
\label{sec:pagestyle}

\begin{figure*}[ht]
\vskip 0.2in
\begin{center}
\centerline{\includegraphics[width=1.9\columnwidth]{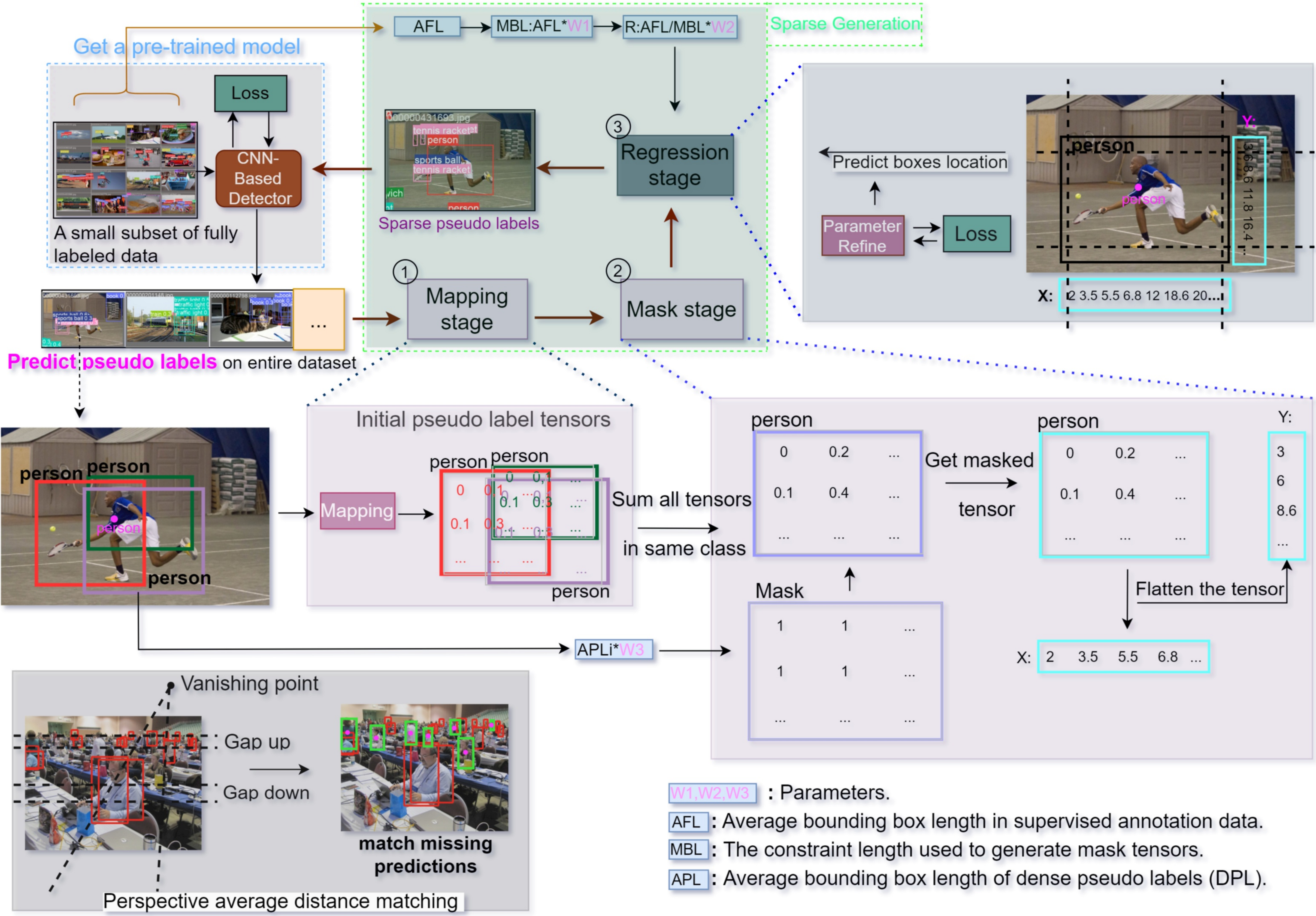}}
\caption{\textbf{The pipeline of Sparse Generation.} It uses non-networked approach and direct regression on pseudo labels.} 
\label{icml-pipeline}
\end{center}
\vskip -0.2in
\end{figure*}

Fig. \ref{icml-pipeline} elaborate the Sparse Generation from its three processing stages. We believe that at low training data volumes, the dense results output and the localized focusing problem can greatly impair the performance. 

For the purpose of making the pseudo labels sparse, at first it is required for efficiently representing the bounding boxes information and distribution, in \textbf{mapping stage}, we propose a simple mapping approach according to design a staircase function and a two-dimensional quadrant. This step is responsible for reflecting the spatial information derived from the dense pseudo labels.

In order to address the issue of localized focusing, \textbf{mask stage} covers as many boxes belonging to the same instance as possible, gets a bounding box map with thermal distribution properties, within the confines of their respective masks. 

In the \textbf{regression stage}, the algorithm computes the final border position of the instance from the previously obtained bounding box map. By regressing it against a small number of GT boxes, this process refine the border position.


\subsection{Mapping stage}

A small amount of supervised annotation data is needed to train an initial model for CNN based detectors to predict pseudo labels, provided that these small amount of supervised annotation data, are followed an overall independent and identically distribution (i.i.d.). These predicted dense pseudo labels (DPL) are mapped into tensors, we hope to reflect the size of each pseudo box on its tensor. Each tensor is designed with a larger value in the center and a smaller value closer to the edge, in order to reduce the impact of some super large span pseudo boxes. These pseudo labels were mapped into tensors through a staircase function $S(x_i,y_i,l_i)$:

\begin{equation}
S(x_i,y_i,l_i) =
\begin{cases} 
0 & \text{if } d_i/l_i > 1 \\
0.1 & \text{if } 0.75 < d_i/l_i \leq 1 \\
0.3 & \text{if } 0.5 < d_i/l_i \leq 0.75 \\
0.6 & \text{if } 0.25 < d_i/l_i \leq 0.5 \\
0.8 & \text{if } 0 < d_i/l_i \leq 0.25 \\
1 & \text{if } d_i = 0  \\
& d_i = \sqrt{(x_i - x_c)^2 + (y_i - y_c)^2}
\end{cases},
\end{equation}

The two-dimensional tensor $IT$ is defined by:

\begin{equation}
IT = 
\begin{cases}
\{ S(x_i,y_i,l_i)\,|\, l_i = w_i \}, & \text{if } w_i \leq h_i\\
\{ S(x_i,y_i,l_i)\,|\, l_i = h_i \}, & \text{if } w_i > h_i\\
\end{cases},
\end{equation}

Where $(x_i,y_i)$ represents the position of each pixel within the pseudo box, $(x_c,y_c)$ represents the center position of the pseudo box, $d_i$ represents the distance between each pixel and pseudo box's center. For saving computational consumption, the tensor $IT$'s size, specifically $h_i$ and $w_i$, are reduced to the half of the size of pseudo box.

\subsection{Mask stage}

In order to reflect the relative position of each pseudo box in the feature space, zero-padding operations are performed on each obtained $IT$ at the scale of entire image.  


Summing all padded tensors $IT'$ with a quantity of $n$ in same class from a picture, to obtain a tensor $ST$ that can reflect the thermal distribution \cite{zhou2019objects} for dense pseudo boxes:

\begin{equation}
ST=\sum_0^{n}IT',
\end{equation}


To avoid interference between pseudo boxes on different instances, covering the tensor $ST$ with a mask tensor $MT$. The size of the mask tensor is determined by the average pseudo boxes length $APL$ in same class of a single picture, which need to ensure that it is bigger than the height and width of the instances in processed images, The Parameter $w3$ is responsible for controlling the size $(w,h)$ of $MT$. For every element $MT(i,j)$ in tensor $MT$ :

\begin{equation}
MT=\{MT(i,j)= 1|i=1,2,...,w,j=1,2,...,h\},
\end{equation}


The obtained mask tensor $MT$ will be padded to the size corresponding to the entire image scale using the coordinates ($x_i$, $y_i$) annotated with point supervision. The padded tensor $MT'$ will be Hadamard product with each summed tensor $ST$ to obtain the tensor after mask coverage $AMT$:

\begin{equation}
AMT= MT'_i \odot ST_i,
\end{equation}

Flattening each tensor $AMT$ onto two one-dimensional tensors, $M_x,M_y$, respectively. Then flattening the tensor into a 0-dimensional tensor \textit{M}. Using the tensor \textit{M} and $M_x$, $M_y$ to obtain \textit{x} and \textit{y} coordinates of their centroids $x_i,y_i$. For every element $AMT_{i,j}, M_{xi}, M_{yi}$ in $AMT$, $M_x, M_y$:

\begin{equation}
\begin{cases} 
M_x=[\sum_0^n AMT_{1,i} \space \sum_0^nAMT_{2,i}...\space  \sum_0^nAMT_{m,i}] \\
M_y=[\sum_0^m AMT_{i,1} \space \sum_0^mAMT_{i,2}...\space  \sum_0^mAMT_{i,n}] 
\end{cases},
\end{equation}


\begin{equation}
M=\sum_0^m M_{xi} + \sum_0^n M_{yi},
\end{equation}

\begin{equation}
\begin{cases}
x_i=\frac{\sum_0^nM_x(j)}{M/2}, \\
y_i=\frac{\sum_0^nM_y(j)}{M/2},
\end{cases}
\end{equation}



\subsection{Regression Stage}

Using the parameter R as a percentage, the predict bounding box location (PBL) function intercepts the position of the border on the x-axis and y-axis on the $M_x$ and $M_y$ tensor, respectively. Then, the information of point annotation is integrated with the PBL predicted result to obtain the pseudo label for a single instance. Subsequently, the algorithm will optimize the parameters based on a limited quantity of supervised annotation data. Its loss function is defined by:

\begin{equation}
Loss=\tanh(\sum_0^n{(SPL_i-GT_j)/n)},
\end{equation}

where \textit{n} is the number of supervised annotation data, $SPL_i$ is the $i$-th sparse label, and $GT_j$ is the corresponding $j$-th supervised annotation data.

\subsection{Perspective average distance matching}

While the volume of labeled data is limited, some instances may not match any corresponding pseudo box predicted by detector, which harm the performance. We observed that the distribution of bounding-box's size in actual images follows the laws of perspective. The existed pseudo boxes could guide the box matching. The method splits the pseudo boxes into two sets (gap up area, gap down area). Calculating the average size of pseudo boxes $ave_{gu}$, $ave_{gd}$ in two gap areas with a distance $d$, use the perspective principle to match the width or height $lenth_i$ of pseudo boxes at each point labeled position $y_i$:



\begin{equation}
lenth_i = (y_i - y_{gu})((ave_{gd} - ave_{gu}) / d) + ave_{gd},
\end{equation}


\section{Experiments}
\label{Experiments}

\subsection{Dataset and experiment detail}

For comprehensive evaluating the performance of Sparse Generation on low data volume, we chosen four datasets: MS COCO-val set (5000 images) \cite{lin2014microsoft}, RSOD \cite{li2020object}, SIMD \cite{haroon2020multisized} and a self built Bullet-Hole dataset.

\textbf{Bullet-Hole dataset}  Due to the need to obtain instances as dense as possible, we conducted Bullet-Hole dataset collection in a real shooting range. In these photos with the highest number of over 502 bullet holes in a single picture.

The split ratio of the training set to the validation set is 10:1 and have been randomly sampled. The classical yoloV5s was selected as the CNN-based detector. Pre-trained model was
obtained by training randomly selected images with a single
card RTX4070 GPU. In the course of conducting experiments on the COCO-val \cite{lin2014microsoft} dataset, no pre-trained weights were employed in the training of the pre-trained model. YoloV5s was trained using pseudo labels generated from each method, all experimental results were obtained after the mAP metric was no longer increased. The P2BNet \cite{chen2022point}, Group Rcnn \cite{zhang2022group} and PLUG \cite{he2023learning}were trained using their official setting with 12, 24 and 12 epochs respectively; the Sparse Generation was trained only 1 epoch. 

\subsection{Experiment Results on MS COCO, Bullet-Hole, RSOD and SIMD Datasets}

\begin{figure}
  \centering
  \includegraphics[width=\columnwidth]{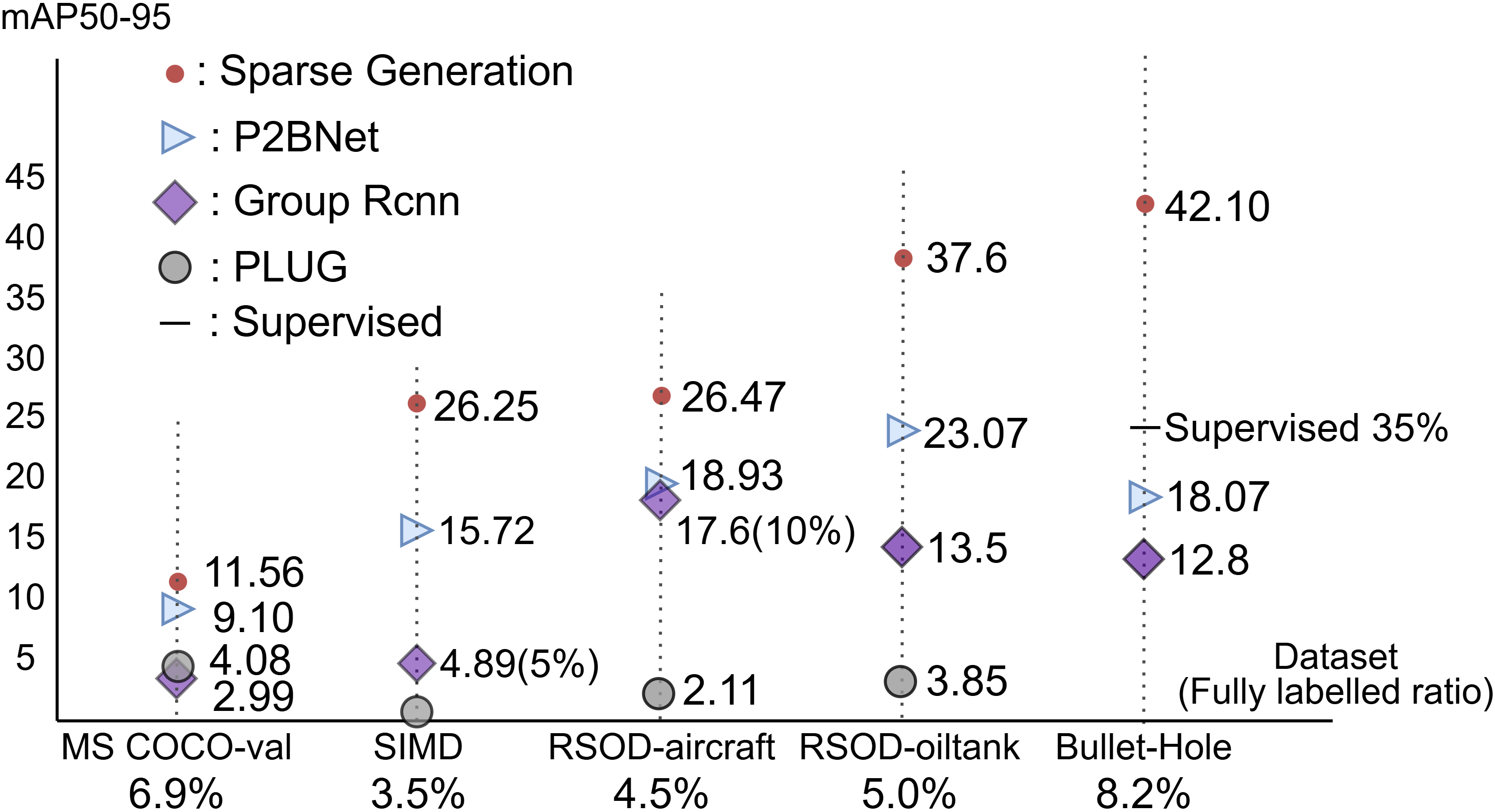} 
  \caption{Experiment result on MS COCO-val \cite{lin2014microsoft}, SIMD \cite{haroon2020multisized}, RSOD \cite{li2020object} and Bullet-Hole datasets.}
  \label{result}
\end{figure}

\begin{table}[t]
\caption{Running time of Sparse Generation on entire MS COCO \cite{lin2014microsoft} dataset.}
\label{table 1}
\begin{center}
\begin{tabular}{cccc}
\toprule
Method  & Time & Device \\ 
\midrule
Sparse Generation    & 16hours & Ryzen5500 cpu \\
P2BNet \cite{chen2022point}   & 48hours+ & RTX4070 gpu \\
Group RCNN \cite{zhang2022group}   & 24hours+ & RTX4070 gpu \\
\bottomrule
\end{tabular}
\end{center}
\end{table}

Fig. \ref{result} depicts the experimental results of Sparse Generation, which evinces advantages over SOTA method at varying labeling ratios. Table \ref{table 1} demonstrates the runtime of Sparse Generation, wherein the non-networked approach renders it more expeditious than other methods, even when executed on CPU. Table \ref{ablation} (a) demonstrates the impact on the results of the distribution of the point labeling distance from the box center. The performance degradation occurred when the point labeling distance from the center point is at 40\%. Table \ref{ablation} (b) illustrates the impact of employing perspective average distance matching (PADM) on the training outcomes. The utilization of PADM has been demonstrated to enhance the detection efficacy, in comparison to the utilization of the average size of pseudo boxes (APL) in an image as a matching approach for undetected instances. Table \ref{ablation} (c) illustrates the impact of the parameter R on the outcomes. The optimal value of the parameter R is relatively modest for the dataset that is not instance-dense (MS COCO), while the optimal value of the parameter R is around 0.1 for the dataset that is instance-dense.
In Table \ref{ablation} (d), 
did not use tensor flattening and calculation of predict boxes location (PBL) function, the biggest bounding box in single instance after mask filtering was combined into one pseudo box, also without the regression of parameters, the performance showed a significant decline.



\begin{table}[tb!]
\caption{Ablation study.}
\label{ablation}
\vskip 0.15in
\begin{center}
\begin{tabular}{cc}

\begin{minipage}[htb]{0.4\linewidth}
    \scalebox{0.85}{
    \centering
    \setlength{\tabcolsep}{3pt}
    \renewcommand\arraystretch{0.65}
    \begin{tabular}{c|c|c}
    \specialrule{0.13em}{0pt}{1pt}
    Range & mAP50 & mAP50-95 \\
    \midrule
    0\% & 63.8 & 25.1 \\
    20\% & 65.9 & 26.5 \\
    40\% & 60.1 & 24.5 \\
    \specialrule{0.13em}{0pt}{0pt}
    \end{tabular}
    }
    \flushleft{\parbox[t]{\linewidth}{(a) The impact of the range of point labels from the center point of the box, on RSOD-aircraft \cite{li2020object}.}}
    \label{tab}
\end{minipage} &

\begin{minipage}[htb]{0.4\linewidth}
    \scalebox{0.85}{
    \centering
    \setlength{\tabcolsep}{3pt}
    \renewcommand\arraystretch{0.65}
    \begin{tabular}{c|c|c|c}
    \specialrule{0.13em}{0pt}{1pt}
    APL & PADM & Data & mAP50 \\
    \midrule
    \checkmark & - & 6.9\% & 30.5 \\
    - & \checkmark & 6.9\% & 31.7 \\
    \specialrule{0.13em}{0pt}{0pt}
    \end{tabular}
    }
    \flushleft{\parbox[t]{1.15\linewidth}{(b) The effect of Perspective Average Distance Matching (PADM), on MS COCO-val set \cite{lin2014microsoft}.}}
    \label{table_17}
\end{minipage}
\end{tabular}\\

\vspace{0.25in} 

\begin{tabular}{cc}
\begin{minipage}[htb]{0.4\linewidth}
    \scalebox{0.8}{
    \centering
    \setlength{\tabcolsep}{3pt}
    \renewcommand\arraystretch{0.65}
    \begin{tabular}{c|c|c}
    \specialrule{0.13em}{0pt}{1pt}
    R  & mAP50 & mAP50-95 \\ 
    \midrule
    \multicolumn{3}{l}{\textbf{\emph{MS COCO-val set \cite{lin2014microsoft}}}} \\
    \midrule
    0.01  & 32.3 & 12.7 \\
    0.03   & 31.8 & 12.0 \\
    0.05    & 31.7 & 12.0 \\
    0.1    & 29.9 & 11.0 \\
    0.2    & 26.5 & 9.15 \\
    \midrule
    \multicolumn{3}{l}{\textbf{\emph{RSOD aircraft dataset \cite{li2020object}}}} \\
    \midrule
    0.01  & 57.6 & 19.7 \\
    0.1  & 63.8 & 25.1 \\
    0.115  & 64.7 & 26.8 \\
    \specialrule{0.13em}{0pt}{0pt}
    \end{tabular}
    }
    \flushleft{\parbox[t]{1\linewidth}{(c) The effect of parameter R.}}
    \label{table_14}
    \end{minipage} \hspace{0.2cm} 

    \begin{minipage}[htb]{0.4\linewidth}
    \scalebox{0.8}{
    \centering
    \setlength{\tabcolsep}{3pt}
    \renewcommand\arraystretch{0.65}
    \begin{tabular}{c|c|c}
    \toprule
     PBL & Parameter Refine  & mAP50  \\ 
    \midrule
    \textbf{-}    & \textbf{-}  & 23.33  \\
    \checkmark    & \textbf{-}  & 83.68  \\
    \checkmark    & \checkmark  & 89.72  \\
    \bottomrule
    \end{tabular}
    }
    \flushleft{\parbox[t]{1.2\linewidth}{\fontfamily{ptm}\selectfont (d) On the RSOD oil tank \cite{li2020object} dataset, remove the function PBL and parameter refine parts, conduct experiments on their functionality.}}
    \label{table_18}
\end{minipage}

\end{tabular}

\end{center}
\vskip -0.1in
\end{table}

\section{Conclusion}

In this paper, we analyzed the shortcomings of previous methods on limited data volume and point weakly supervised object detection for dense instances tasks. Using CNN-based network specially for pseudo labels generation, the output results will still be a relatively dense set, and prone to localized focusing problem. The use of these pseudo labels subsets as a supervision for model training will result in a reduction in the model's performance. A non-networked pseudo label sparsity method was proposed, which has only three parameters, achieves better performance for training only one epoch.


\vfill\pagebreak

\bibliographystyle{IEEEbib}
\bibliography{strings}
\end{document}